# Universal and Transferable Attacks on Pathology Foundation Models

Yuntian Wang[1,2,3†], Xilin Yang[1,2,3†], Che-Yung Shen[1,2,3], Nir Pillar[4] and Aydogan Ozcan[1,2,3,5*]

[1]Electrical and Computer Engineering Department, University of California, Los Angeles, CA, 90095, USA

[2]Bioengineering Department, University of California, Los Angeles, CA, 90095, USA

[3]California NanoSystems Institute (CNSI), University of California, Los Angeles, CA, 90095, USA

[4]Department of Pathology, Hadassah Hebrew University Medical Center, Jerusalem, 91120, Israel

[5]Department of Surgery, University of California, Los Angeles, CA, 90095, USA

[†]These authors contributed equally to the work

[*]Correspondence to: ozcan@ucla.edu

## Abstract

The advent of foundation models has initiated a paradigm shift in digital pathology with a plethora of emerging opportunities. However, these powerful systems also introduce centralized vulnerabilities, making them critically susceptible to sophisticated adversarial attacks that might pose a significant threat to patient safety. To shed light on these potential threats, here we introduce **U**niversal and **T**ransferable **A**dversarial **P**erturbations (**UTAP**) for pathology foundation models that reveal critical vulnerabilities in their capabilities. Optimized using deep learning, UTAP comprises a fixed and weak noise pattern that, when added to a pathology image, systematically disrupts the feature representation capabilities of multiple pathology foundation models. Therefore, UTAP induces performance drops in downstream tasks that utilize foundation models, including misclassification across a wide range of unseen data distributions. In addition to compromising the model performance, we demonstrate two key features of UTAP: (1) *universality*: its perturbation can be applied across diverse field-of-views (FOV) independent of the dataset that UTAP was developed on, and (2) *transferability:* its perturbation can successfully degrade the performance of various external, black-box pathology foundation models – never seen before. These two features indicate that UTAP is not a dedicated attack associated with a specific foundation model or image dataset, but rather constitutes a broad threat to various emerging pathology foundation models and their applications. We systematically evaluated UTAP across various state-of-the-art pathology foundation models on multiple datasets, causing a significant drop in their performance with visually imperceptible modifications to the input images using a fixed noise pattern. The development of these potent attacks establishes a critical, high-standard benchmark for model robustness evaluation, highlighting a need for advancing defense mechanisms and potentially providing the necessary assets for adversarial training to ensure the safe and reliable deployment of AI in pathology.

## Introduction

The integration of Artificial Intelligence (AI) into computational pathology, utilizing general-purpose foundation models, has significantly enhanced performance on a vast range of tasks, including cancer detection, subtyping, grading, and staging[1–6], compared to task-expert AI models that specialize in only one task. Such foundation models are predominantly self-supervised by either contrastive learning or masked modeling[7–11] and effectively utilize Vision Transformers (ViT)[12] as their backbone. These foundation models are pre-trained at large scale using e.g., millions of pathology slides, and then used as feature extractors for diverse downstream tasks[13–17]. The self-attention mechanism—central to ViTs —enables long-range context, but its global token coupling can also be exploited for attacks: attention-aware or patch-based adversarial methods have shown effective targeted perturbations against ViTs[18–25]. These adversarial attacks, utilizing subtle and carefully crafted perturbations, are often imperceptible to human observers but can induce catastrophic model failures[25–27], ranging from representation power collapse to attacker-directed hallucinations. Such an attack could, for instance, cause a model to misdiagnose a malignant tumor as benign, posing a direct and severe risk to patients, especially without human experts in the loop.

In the domain of adversarial attacks, earlier efforts predominantly focused on a single or a small set of specific image samples or models, resulting in attacks that are highly specialized in scope[18,21]. In this work, we move beyond model- or image patch-specific attacks to explore more broadly applicable threats to pathology foundation models characterized by both *universality* and *transferability*. For exploring the vulnerabilities of pathology-related foundation models, we developed a universal and transferable adversarial perturbation (UTAP) framework, which utilizes an adaptive Projected Gradient Descent (PGD)-based[28] optimization process to iteratively craft a subtle noise pattern. We define universality as the capability of a single, fixed perturbation that applies across a wide range of models and data, including unseen images from different datasets to cause model failure. We define transferability as an attack crafted against one foundation model, where it can successfully compromise, without any adjustments to its perturbation, other foundation models, which are never seen before, i.e., neither the parameters nor the gradients of the attacked models are accessible to the attacker. We improved the universality and transferability of our adversarial attacks with two key designs. First, we iteratively optimize the attack across a set of training data to disrupt the model's output feature space by minimizing the similarity between the feature representations of the original and perturbed/attacked images. This strategy fundamentally hampers the model's representational power and grants the perturbations superior universality across various image patches. Second, we add random masking and attention dropping regularization when computing the gradients to ensure better transferability to other foundation models never seen before, i.e., following a black-box setting. The efficacy of UTAP was comprehensively validated by demonstrating that a single, fixed perturbation successfully degrades the performance of seven state-of-the-art pathology models[29–35], significantly reducing the probing accuracy in downstream classification tasks while simultaneously compromising the representation power of all the tested models.

UTAP shows that one universal perturbation trained in one setting transfers to unseen datasets and other foundation models without any modifications, revealing system-level vulnerabilities that standard evaluation methods can miss. To support safer development and deployment of pathology foundation models, it is essential to comprehensively study these threats and develop robust defenses. Framed as "ethical hacking", our work develops and benchmarks advanced attacks and aims to raise awareness, with the goal of enabling powerful tools needed to build, assess, and improve the next generation of resilient, secure computational pathology models.

## Results

**Universal and Transferable Adversarial Attacks on Pathology Foundation Models**

We first validate our attack strategy on a downstream classification task using a dataset of 100,000 non-overlapping patches from hematoxylin and eosin (H&E) stained histological images of human colorectal cancer (CRC) and normal tissue[35]. As illustrated in **Fig. 1** and in the Methods section, the attack generation scheme involves iteratively optimizing a single, trainable UTAP pattern via the adaptive PGD[28] method with random masking and attention dropping regularization[36]. For details of the implementation refer to the Methods section. During the training, the UTAP-generated noise pattern is added to original images, and both the original and the attacked images are passed through a frozen pathology foundation model to extract their respective feature representations. The core of our optimization aim is to minimize the cosine similarity between the original and attacked feature embeddings, thereby forcing them to be as dissimilar as possible in the feature space, corrupting the representation power of the foundation model. Once the training converged and the perturbation is finalized, the UTAP's effectiveness was comprehensively evaluated using a methodology to assess universality and transferability concurrently. To assess the performance of classification downstream tasks, we first trained a light-weight linear classifier on the extracted features from clean images (classification tokens / CLS tokens) in the classification training dataset and then computed the blind testing classification accuracy on the corresponding test dataset; this sets the baseline accuracy in each task before the attack (see Methods for details). We then applied a single, fixed UTAP perturbation pattern to the images in the same testing dataset that neither the classifier nor UTAP saw during their optimization. These perturbed/attacked images were then evaluated against multiple foundation models, including the original model used during UTAP optimization (internal generalization) and new, unseen foundation models (external generalization). The attack's success was quantified by the drop in the classification accuracy in each case—computed as the difference between the original and attacked accuracy values—using the well optimized pre-trained linear classifier for each foundation model and data combination. This testing pipeline allows us to confirm the attack's universality by quantifying its impact on new, unseen data, and its transferability by assessing the performance degradation across different foundation models – never seen before.

The efficacy of the UTAP attack was qualitatively and quantitatively evaluated in **Fig. 2**. The perturbation pattern itself, visualized against a neutral gray background in **Fig. 2a**, is composed of pixel values strictly bounded within a small range of $[-\epsilon, \epsilon]$. As a direct consequence of this constraint, when the UTAP perturbation is added to the original images, the resulting attacked images are visually almost indistinguishable from their unperturbed counterparts. This perceptual subtlety is also demonstrated in **Fig. 2a** with a side-by-side comparison of the original and attacked tissue patches from three distinct classes: adipose (ADI), lymphocyte (LYM), and normal colon mucosa (NORM). Further analysis of UTAP's statistical properties is reported in **Fig. 2b**, which displays a histogram of its pixel-wise values averaged from the three-color channels. The histogram plot reveals a distribution centered approximately at zero and exhibits prominent spikes at the $\pm\epsilon$ bounds, a characteristic outcome of the hard energy constraints enforced by the adaptive PGD optimization algorithm used to generate the attack (see the Methods section).

While UTAP is designed to be visually imperceptible, its impact on model performance is substantial. The impact of the attack on classification accuracy was quantitatively reported in **Fig. 2c**, which details the attack's performance across seven different foundation models while we only used one foundation model (UNI2-h) for training UTAP. The chart compares the original classification accuracy of each model (green bars) with the accuracy on the images attacked by UTAP (blue bars). The substantial drop in performance, quantified for each model (red bars), demonstrates the attack's ability to induce significant model failure. Critically, this degradation is observed not only in the model used to train the perturbation (UNI2-h in this case) but across all the other six external, black-box foundation models (the weights and gradients of which were *not* known), confirming the attack's high degree of transferability. It is important to note that this accuracy drop is not caused by an attack on the downstream linear probing layer, but rather stems directly from the corruption of the upstream feature representations, which is also confirmed with the analysis in **Fig. 2d-e**. To shed more light on our attack's impact on the feature space representation, **Fig. 2d** displays the Uniform Manifold Approximation Projections (UMAP)[37] of the feature embedding of image patches extracted by the foundation model before and after the attack. The left and middle scatter plots show that in the absence of an attack, the training and testing feature embeddings of different tissue classes form distinct, well-separated clusters through UMAP. In contrast, the right scatter plot reveals that after applying the optimized UTAP perturbation to the test images, these orderly clusters collapse into chaotic and largely undifferentiated structures. This dimensionality reduction confirms that the UTAP operates by fundamentally corrupting the foundation model's feature representation capability, causing the feature space to collapse, destroying the linear separability of features, and thereby crippling its discriminative classification performance. **Figure 2e** further illustrates this effect through the Principal Component Analysis (PCA) of the feature representations in addition to UMAP. These visualization results confirm a similar insight that the attacked features undergo a mode collapse, forming a tight, squeezed cluster in contrast to the well-separated original features. These analyses demonstrated that UTAP was able to not only cause significant classification accuracy drops, but also comprise the general representation power of the foundation models, even for black-box foundation models that were never seen before.

To further understand the effect of the UTAP attack and its behavior, we compared the probabilities predicted by the linear classifier from the original image data versus the probabilities resulting from the attacked images. **Supplementary Fig. S1a** presents the results of the attack on UNI2-h model alongside the corresponding original and attacked images shown on the left for all nine data classes. The results highlight a key difference between UTAP and traditional adversarial attack methods. Conventional attacks are typically targeted, i.e., they have a predefined attack goal, manipulating inputs to maximize the probability for a specific incorrect class[28,38]. In contrast, as shown in **Supplementary Fig. S1b**, UTAP focuses on degrading the feature space itself, without a fixed directional preference, and therefore it does not manifest itself as a consistent peak for any single incorrect class, but rather generally corrupts the model's confidence, resulting in a more uniform distribution across all false classes. This non-targeted behavior is further demonstrated in **Supplementary Fig. S2**, which compares the classification probability distributions for the internal foundation model (UNI2-h) against several external foundation models (e.g., Gigapath and Virchow2) which were not used during the training phase. The figure reveals that while the attack degrades performance across all the foundation models, the specific characteristics of the resulting probability distributions are model-dependent. These variations on the predicted probability distributions underscore that UTAP does not force a single, predictable outcome. Instead, it interacts with each foundation model with a unique behavior, providing further evidence that it operates by fundamentally disrupting the feature representation capabilities with high transferability to black-box foundation models, never seen before.

We further investigated UTAP attack's impact on the model's spatial features by visualizing the cosine similarity between the classification/[CLS] tokens and all the local visual tokens (see the Methods) and plotting them as a heatmap in the corresponding locations on the whole slide image. As illustrated in **Fig. 3**, before the attack, the resulting heatmaps clearly show the foundation model's focus on diagnostically relevant tissue structures, indicated by high similarity (red hues), while assigning low similarity values to irrelevant regions (blue hues). Following the application of UTAP perturbation, the attacked attention maps become profoundly disrupted, exhibiting a scattered and incoherent distribution of cosine similarity values. This degradation provides direct evidence that our attack does not merely manipulate a global decision boundary but fundamentally impairs the model's core ability to extract meaningful and localized features across the entire image, thereby compromising its foundational understanding and representation of the tissue content.

To better understand the impact of the regularization method used in UTAP training, we also conducted an ablation study evaluating the efficacy of UTAP trained *with* and *without* attention dropping and random masking. The results reported in **Supplementary Fig. S3** reveal an intuitive trade-off: removing the regularization techniques boosts the attack's efficacy against the training foundation model (resulting in improved internal attacks) but significantly compromises its transferability to external models (resulting in worse external attacks). This is because unregularized perturbations overfit to the training model by exploiting model-specific attention patterns. As these patterns are not strictly conserved across different foundation models, the perturbation fails to generalize, diminishing its effectiveness in black-box attacks to other foundation models, never seen before.

To further assess the universality of UTAP across different datasets, we evaluated the attack's impact on an entirely unseen, out-of-distribution dataset. We assembled an external test set by selecting six classes from The Cancer Genome Atlas Program (TCGA) Uniform Tumor dataset[39], none of which were used during the optimization process. Applied without any adaptation, the pre-optimized UTAP pattern produced ~20% drop in the classification accuracy across all the evaluated models (see **Supplementary Fig. S4**). This successful attack on out-of-distribution (OOD) images further demonstrate the universality of UTAP perturbation method, indicating that UTAP exploits dataset-agnostic, feature-level vulnerabilities, underscoring a broader security risk beyond the source distribution/dataset.

To further quantify transferability of UTAP, we conducted a systematic cross-model attack analysis,

summarized in **Fig. 4**. UTAP optimization in this analysis used the same intensity constraint and step schedule, and evaluation is performed on unseen patches for every model; see the Methods for details. This evaluation is structured as a matrix-like representation where the row indexes represent the foundation model used to train the UTAP perturbation that is used, while the column indexes represent the foundation model being attacked and evaluated. The diagonal entries, highlighted by the dashed green boxes, correspond to internal attacks, where the perturbation is evaluated on the same foundation model architecture it was developed on. The off-diagonal elements represent external attacks, simulating a black-box scenario where a perturbation trained on one foundation model is directly used to attack other unseen foundation models. Crucially, all attacks presented in this analysis were performed on new, unseen image patches. The results of this analysis reveal two critical findings. First, the internal attacks show that the UTAP perturbation process is consistently powerful, reducing accuracy on the models on which it was trained. More importantly, the external attack results reveal that UTAP exhibits a high degree of transferability across all training-testing model pairs. For instance, a UTAP perturbation trained on UNI2-h not only reduces its own accuracy down to 12.27% but also successfully degrades the performance of other foundation models, such as lowering Prov-Gigapath's accuracy from 96.42 to 48.69% and Virchow2's accuracy from 97.23% to 25.52%.

We further evaluated the transferability of UTAP by training it with multi-source optimization, in which the source model is randomly switched every eight steps from a fix-sized pool drawn from seven foundation models (detailed in **Supplementary Fig. S5**). All UTAP optimization steps used the same intensity constraint and step schedule, and evaluation is performed on unseen image patches for every model. In each optimization process, the foundation models included in the pool are treated as internal (dashed green box), while the excluded models are treated as external (orange box). The results reveal two key trends. First, the foundation models present in the training pool experience markedly larger accuracy drops than external models, indicating that direct exposure to the foundation model during UTAP training still confers an advantage to the attack. Second, as the pool size increases, transfer to external foundation models becomes broader and more uniform in their attack performance, but the per-model impact on internal targets relatively diminishes. Intuitively, optimizing the attack against a diverse group of foundation models steers the perturbation toward features shared across these models (which improves transfer success), while conflicting gradients prevent over-specialization to any single model weight or architecture (reducing peak severity). Practically, this suggests that an attacker can tune the size of the model pool to balance maximal depth on a specific model versus wide, cross-model disruption. With these analyses, we demonstrated the ability of UTAP to significantly degrade the performances of multiple, unseen external foundation models; these results underscore a critical vulnerability, highlighting a need for advanced defense mechanisms that are robust against not only direct attacks but also these more insidious, transferable threats.

Having established how training diversity modulates transferability, next we examine how optimization choices shape UTAP's performance by exploring two key hyperparameters: $\theta$ which is the scaling factor for the step size and $\epsilon$ which is the allowable magnitude of the perturbation; see **Fig. 5**. We first analyzed the impact of $\theta$, by training all UTAP perturbations on the UNI2-h model and evaluating them across all seven foundation models while varying $\theta$ ($\epsilon$ fixed). The histogram of the UTAP perturbations and the attacked images for different choices of $\theta$ are visualized in **Fig. 5a**. In these results, we observe that larger $\theta$ drives the perturbation distribution to become more bimodal, concentrating at the $\pm\epsilon$ boundaries. The quantitative accuracy evaluations reported in **Fig. 5b** reveal a distinct difference between the internal and external attack performances. The internal attack efficacy on UNI2-h increases monotonically with $\theta$, whereas the external attack efficacy is non-monotonic: the highest accuracy drop is observed at an intermediate $\theta = 10$ and then weakens. Thus, step size critically governs transferability, with an optimal $\theta$ value balancing internal strength and cross-model generalization of the attack. Similarly, with $\theta$ being fixed, we also analyzed the impact of $\epsilon$—the perturbation bound—which sets the maximum magnitude of the perturbation pattern. The analyses and visualizations reported in **Fig. 5c** reveal that as $\epsilon$ increases, the perturbation becomes stronger and as anticipated, the attack gets visually noticeable on the resulting images. This increase in perturbation strength correlates directly with attack efficacy, as shown in **Fig. 5d**: accuracy drops due the attack grow monotonically with increasing $\epsilon$ across all the targeted models. Consequently, $\epsilon$ governs the potency–stealth trade-off: larger bounds yield stronger but more perceptible attacks.

**Patch-Specific and Class-Specific Adversarial Perturbations**

Next we consider image patch-specific and data class-specific adversarial perturbations. Consider a scenario in which an adversary targets a specific patient whose tissue specimen is submitted for diagnostic analysis. Depending on the adversary's prior knowledge of the patient, the required universality of the attack varies along a hierarchy from the most to the least specific. At the most specific level (low universality), if the attacker obtains an exact digital copy of the patient's tissue patch, they can craft a bespoke perturbation; we term this attack as **P**atch-**S**pecific **A**dversarial **P**erturbation (PSAP), which optimizes a unique perturbation pattern per image patch for an adversarial attack. With only class-level knowledge (e.g., the specimen's diagnostic category), the attacker can also deploy a class-conditional perturbation; we model this scenario using the **C**lass-**S**pecific **A**dversarial **P**erturbation (CSAP) to cause misclassification of a specific data class. At the least specific/most universal strategy, an attacker possesses no prior knowledge of the specific image or its class label, necessitating a universal attack (UTAP) that can degrade model performance across all inputs and models – as we reported in our former sub-section. At the lowest level of universality, we evaluate image patch-specific adversarial perturbations (shown in **Fig. 6**), where an attacker has knowledge of a targeted, specific sample image. In this attack strategy, each image patch receives a trainable perturbation, which is added to the corresponding patch before the attacked image is processed by the frozen pathology foundation model and its pre-trained linear classifier. To induce misclassifications, the perturbations are iteratively optimized by maximizing the cross-entropy loss between the model's prediction and the one-hot ground-truth label (detailed in Methods section). This attack strategy reveals a critical trade-off between internal attack performance and external generalization: tailored to individual image patches, PSAP collapses the classification accuracy on the optimized samples from 97.37% to 0.00% where none of the predictions are correct. However, the attack fails in terms of universality, leaving performance on unseen images largely intact (95.00 % classification accuracy). Thus, while patch-specific attacks can severely compromise a model on specific image data, their generalization capability is limited since the attack design is patch-specific.

Next, we evaluated class-specific adversarial perturbations that target all images within a single tissue class (see **Fig. 7**). A single perturbation was trained per class—for example, one fixed perturbation pattern for all adipose (ADI) images and another one for all lymphocyte (LYM) images—and then added to each image in that data class. The perturbated/attacked images were then passed through the frozen pathology foundation model with its pre-trained linear classifier. Similar to the perturbation training method used in PSAP, the optimization of CSAP maximizes the cross-entropy between the model's prediction and the one-hot ground-truth label over the class dataset (see the Methods). CSAP is highly effective within its desired scope, reducing the accuracy on its targeted class from 97.37% to 1.42% on training images and to 29.80% on unseen test images. Therefore, CSAP generalizes within its attacked data class, and yet remains specialized, showing little effect on other tissue types/classes due its design.

Compared to PSAP and CSAP, UTAP presents, by its design, the highest level of universality—a single, image-agnostic perturbation pattern applicable to almost all unseen samples. As summarized in **Fig. 8**, UTAP shows the strongest and most consistent generalization, yielding comparable degradation of classification accuracy on training (14.10%) and unseen test images (12.27%). Overall, as universality increases, overfitting diminishes and the attack's impact on novel images strengthens, making UTAP a practical and consequential threat.

## Discussion

We presented UTAP, a single, universal, perturbation that transfers across different pathology foundation models. In contrast to logit-level attacks that maximize cross-entropy to induce class flips along model-specific decision boundaries, UTAP intervenes earlier in the pipeline by directly disrupting the feature representations. Our training scheme is designed to disrupt a model's core feature representations on classification/[CLS] tokens and also local patch features, yielding perturbations that are highly transferable and capable of compromising black-box foundation models on unseen images, *without* any model-specific or instance-specific post-training.

A central advantage of UTAP is its strong cross-model transferability demonstrated across various foundation models. UTAP targets the representation itself, corrupting a more fundamental and widely shared structure. During UTAP training, we optimize the perturbation using a cosine-similarity loss—rather than the cross-entropy objective used in PSAP and CSAP—to directly decorrelate the attacked feature vectors from the original representations. This objective drives the features to become maximally dissimilar, and compels the optimization to exploit the full spectrum of activations, including negative values, thereby reorienting the vectors in feature space. By manipulating the geometry of the representation manifold instead of merely disrupting logits across a decision boundary, UTAP induces a more fundamental corruption of representational power, which likely underpins its superior transferability across architectures and different black-box models. This effect is also evident in **Fig. 2d–e**: UMAP and PCA projections reveal a collapse of the feature manifold, preventing the formation of semantically meaningful clusters after the UTAP attacks. This strategy is particularly powerful against ViTs, which operate by dividing images into a grid of patches and learning local feature representations for each. By incorporating attention dropping and random masking regularization techniques during the training stage, the UTAP perturbation was forced to focus its optimization on disrupting these local, patch-level features rather than the more complex inter-patch relationships learned by the attention mechanism[36], which resulted in a greater classification accuracy drop of external models as reported in **Supplementary Fig. S3**. The resulting grid-like pattern that is often visible in UTAP arises from its interaction with the ViT's patch-embedding stage, where the perturbation aligns with position token boundaries and interacts with the model's patch-based representation. The fact that this single approach is effective across numerous, independently trained foundation models suggests they have all converged on a similar scheme for representing fundamental histopathological features, making this shared representation a critical and exploitable vulnerability.

To investigate the influence of the training set size on the attack's efficacy, we conducted an ablation study by varying the number of images ($N$) used to optimize UTAP. As shown in **Supplementary Fig. S6**, increasing the number of training images from $N = 900, 1800, 2700$ to $N = 100k$ (full training dataset) yielded no significant improvement in the attack's performance across both internal and external foundation models. Post-attack classification accuracies were largely consistent across settings with small variations. This suggests that a modest, diverse image dataset suffices to capture generalizable vulnerabilities of pathology foundation models in representation space. To conserve computation, we therefore used $N = 900$ in our results unless otherwise noted; the training time for a single UTAP pattern on 900 training samples took ~13 min.

To further confirm that the efficacy of our attacks stems from UTAP's optimized delicate structure rather than the mere addition of noise, we conducted an ablation study using a randomly generated noise perturbation of the same intensity level. For this experiment, we generated a random noise pattern with pixel values uniformly sampled within the same magnitude constraint of $\epsilon = 20$ used for UTAP. As illustrated in **Supplementary Fig. S7**, adding this unstructured random noise to the input images had a negligible effect on the performance of all the tested foundation models. The classification accuracy remained high across the board, with only a minimal drop observed. This result demonstrates that the foundation models are inherently resilient to unstructured, random perturbations of this magnitude. It further underscores that the potency of UTAP is not simply a consequence of its energy or magnitude of the noise added but is derived from its carefully crafted, adversarial structure, which is specifically optimized to disrupt the fundamental feature representations across pathology foundation models.

While this work provides a comprehensive analysis of universal and transferable attacks, our investigation centered on the vulnerabilities of ViT-based foundation models, which are currently the dominant design choice in almost all visual foundation models. We also expect that UTAP can be extended with minor modifications to attack Convolutional Neural Network (CNN)-based models[40] or hybrid models[12,41] which is left as future work.

Our evaluations in this work were focused on the downstream task of tissue classification on the image patch-level. Our approach can be further extended to other pathology tasks, such as semantic segmentation[29,42] and cell counting[43], to understand how these universal perturbations impact models' ability to perform fine-grained spatial analysis and quantification. Broadening the investigation across these dimensions will be essential for developing a truly holistic understanding of AI safety in digital pathology. Furthermore, it is crucial to investigate the impact of adversarial perturbations on WSI (whole slide image)-level tasks. Many slide-level diagnostic models operate with weakly-supervised conditioning and aggregate features from numerous patches in different ways. Most opt to use only feature vectors of their [CLS]/classification tokens through multi-instance learning[45], cascaded transformers[29,44] or other architectures like LongNet[45,46]. We expect that since the proposed attack successfully distorts these fundamental patch-level representations, the performance of any WSI-level model that depends on them will inevitably degrade as well. Extending the presented analysis to these multi-scale tasks will be important for a comprehensive view of AI safety.

The rapid creation of a universal and transferable perturbation pattern, in less than 15 min of training time, carries significant implications for the clinical deployment and safety of AI in pathology. The demonstrated success of a transferable, black-box attack indicates that even proprietary—and potentially FDA-regulated—diagnostic systems can be compromised without access to the model, posing a direct risk to patient safety. This vulnerability highlights the need for mandatory adversarial-robustness evaluations and watchdog development efforts prior to deployment, extending beyond standard accuracy metrics to ensure resilience against sophisticated threats. Crucially, this work provides both a rigorous benchmark and a practical pathway to defense. UTAP serves not only as a stress test for foundation model robustness but also as a training signal for hardening models in practice. By incorporating UTAP-generated perturbations into adversarial training[47,48], developers can effectively "vaccinate" models against a broad class of stealthy, transferable attacks. This strategy can potentially strengthen intrinsic safety, support regulatory confidence, and advance the reliable clinical adoption of AI in digital pathology.

## Methods

**Adversarial Attack Dataset and Pathology Foundation Models**

The primary dataset used in this study is the NCT-CRC-HE-100K (short as CRC-100K)[35], a collection of 100,000 non-overlapping image patches from H&E stained histological images of human colorectal cancer and normal tissue. All images are 224x224 pixels at a resolution of 0.5 μm per pixel with a 20x objective and were color-normalized using the Macenko's[49] method. The dataset comprises nine manually annotated tissue classes: Adipose (ADI), background (BACK), debris (DEB), lymphocytes (LYM), mucus (MUC), smooth muscle (MUS), normal colon mucosa (NORM), cancer-associated stroma (STR), and colorectal adenocarcinoma epithelium (TUM). These images were manually extracted from 86 H&E-stained formalin-fixed paraffin-embedded (FFPE) whole slides from the NCT Biobank and the UMM pathology archive, which included CRC primary tumors, liver metastases, and normal tissue regions from gastrectomy specimens to increase variability. For validation, an independent test set, CRC-VAL-HE-7K[35], was used, containing 7,180 image patches from 50 patients with colorectal adenocarcinoma who were not part of the NCT-CRC-HE-100K (training) cohort. In the further evaluation of UTAP's universality (shown in **Supplementary Fig. S4**) on external datasets, we used the TCGA Uniform Tumor dataset[39], comprising 1,608,060 image patches extracted from H&E-stained histological samples across 32 solid tumor types. To establish a simplified testbed, we restricted the label space to a subset of six representative classes—Adrenocortical, Bladder, Cervical, Esophageal, Kidney, and Mesothelioma. Patches from TCGA Uniform Tumor dataset originally sized 256 × 256 pixels were center cropped to 224 × 224 to match the input dimensions of the pre-optimized UTAP.

We primarily used a foundation model pool containing the following seven pathology foundation models directly accessed from HuggingFace: [*MahmoodLab/UNI2-h, prov-gigapath/prov-gigapath, paige-ai/Virchow2, paige-ai/Virchow, bioptimus/H-optimus-0, bioptimus/H0-mini, bioptimus/H-optimus-1*][29–35]. All the models have the vision transformer architecture as backbone, achieving state-of-the-art performances across multiple clinical downstream tasks in histopathology. A typical ViT forward pass partitions an image into non-overlapping patches, flattens each patch, and projects it with a trainable linear layer to obtain patch embeddings. A learned classification token ([CLS])[50] is first prepended to the sequence and the learned positional embeddings are later added. The sequence then passes through a stack of Transformer encoder blocks, each composed of multi-head self-attention (via query, key, and value projections) and a multiple layer perceptron (MLP), with LayerNorm and residual connections[51]. While our experiments aim at attacking foundation models using ViT architectures with a 14-pixel patch size as default design choice, the approach is broadly applicable to ViTs with different patch encoding schemes, including patch sizes of 16 or 28 pixels.

**Visualization of Adversarial Perturbations**

To effectively visualize the structure of the adversarial perturbation, we superimposed the perturbation onto a uniform gray background (R, G, B values: [128, 128, 128]). This method makes subtle, high-frequency patterns of perturbation visible against a neutral gray image. For histogram analysis, the perturbation's RGB channels were averaged to create a single-channel intensity map. The histogram was then computed from the pixel intensity distribution of this grayscale representation.

**Parameters of Digital Implementation and Training Scheme**

For UTAP training, $N = 900$ image patches of $224 \times 224$ pixels were sampled from the training dataset with 100 images per class. The original and attacked images were normalized using the same methods as the corresponding foundation model was trained with. UTAP was optimized for $L = 10$ epochs iterating by $B = 5$ images per batch. The perturbation was updated using adaptive PGD methods with $\theta = 10$, step size $\alpha = \frac{\epsilon \theta B}{255 \times LN} = 4.35 \times 10^{-4}$ and clamped to the range between $[-\epsilon, \epsilon]$ where $\epsilon = 20$ after each iteration.

The numerical simulations and the training process for UTAP were carried out using Python (version 3.11.15) and PyTorch (version 2.1.2, Meta Platform Inc.). The UTAP optimization underwent a 10-epoch training on a workstation equipped with an Nvidia GeForce RTX 4090 GPU, an Intel Core i9-13900KF CPU, and 32 GB RAM. The training time for a single UTAP ($224 \times 224 \times 3$ pixels) on 900 samples was ~13 minutes.

## Supplementary Information

Supplementary Information includes:

- Supplementary Figures S1-S7
- Adversarial Perturbation Training and Evaluation
- Training Loss Functions and Backend Classifier Training

# Figures

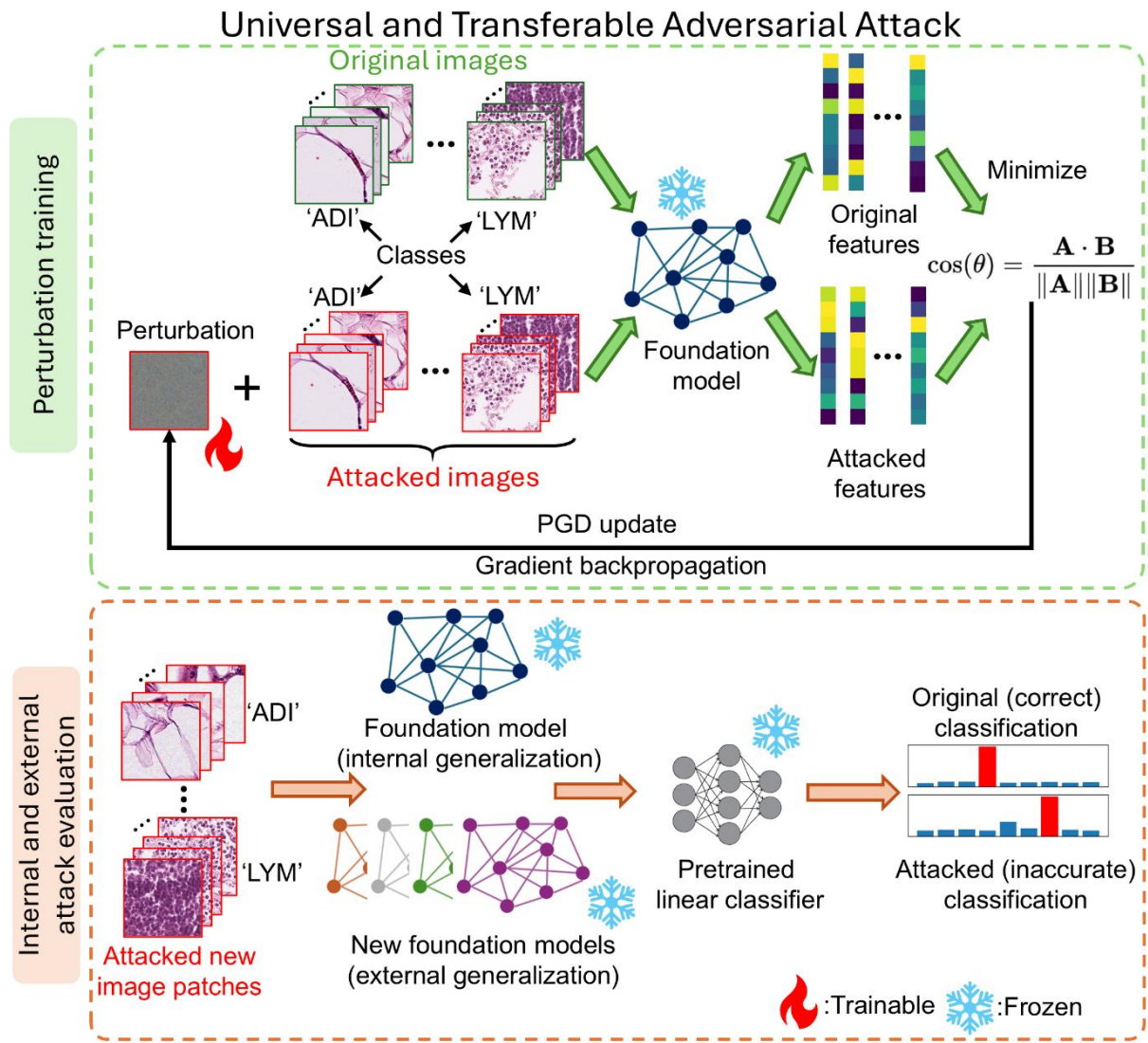

**Figure 1. Workflow of Universal and Transferable Adversarial Perturbation (UTAP) training and evaluation.** Top row shows the process of UTAP training which minimizes the cosine similarity of original and attacked features extracted from a frozen foundation model. The bottom row is the evaluation process of UTAP attack, where the attacked images were passed through a frozen foundation model used for training (internal generalization) and unseen, black-box foundation models (external generalization) cascaded with their respectively pretrained linear classifiers. The fire and snowflake symbols indicate whether the parameters are trainable or frozen/fixed.

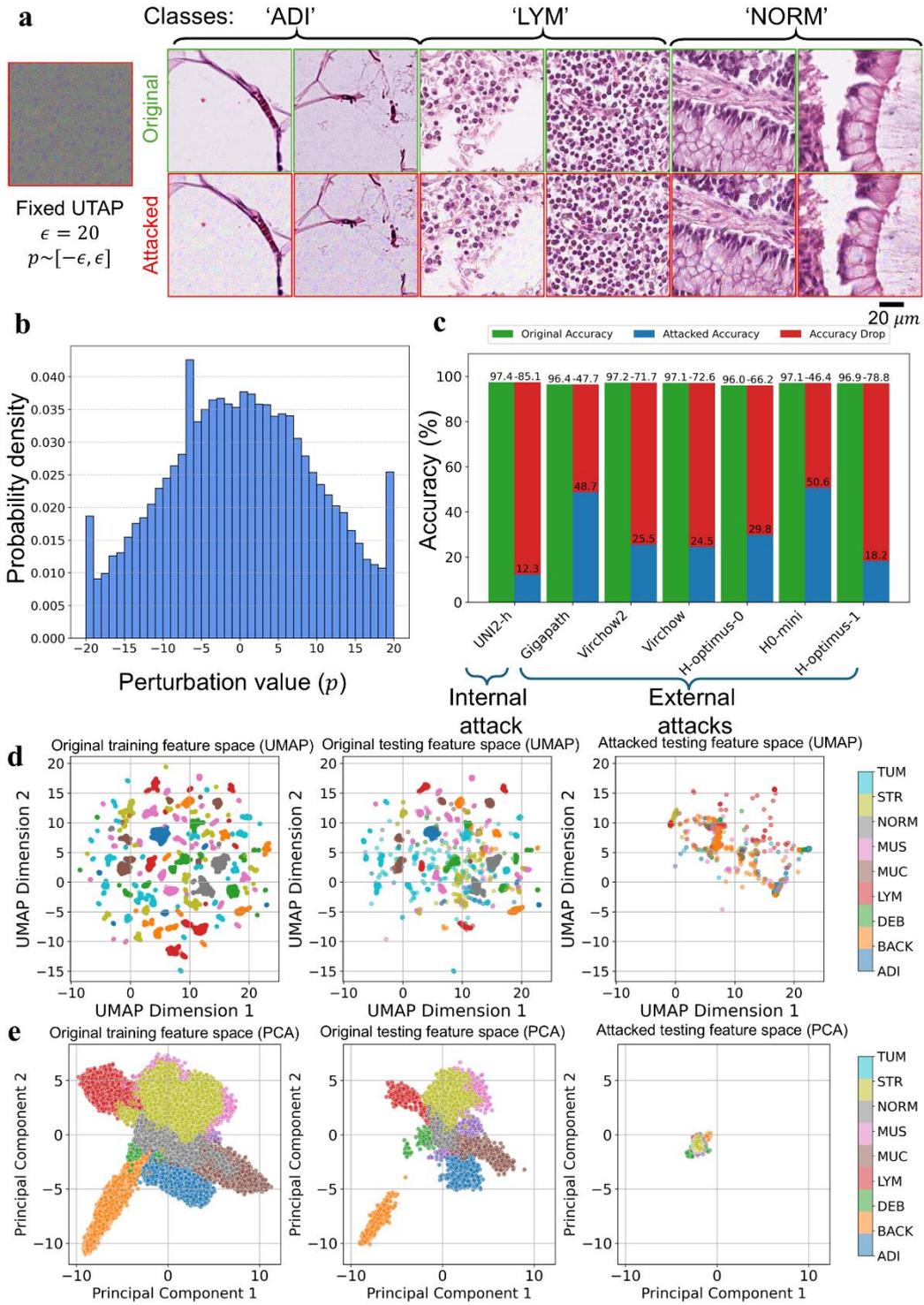

**Figure 2. UTAP attack results on CRC-100K dataset.** (a) Visualization of the optimized UTAP with the original and attacked images of three classes ('ADI' for adipose, 'LYM' for lymphocytes and 'NORM' for normal colon mucosa) sampled from the CRC-100K dataset. (b) The histogram of UTAP perturbation, bounded in the range of $\pm\epsilon$, where $\epsilon = 20$. (c) The original (green), attacked (blue) and dropped (red)

classification accuracy values on the internal attack (the foundation model used for UTAP training) and various external attacks (for unseen foundation models). (d) the UMAP projections into the 2D space of the original features and the attacked features. (e) The PCA projections into the 2D space of the original features and the attacked features.

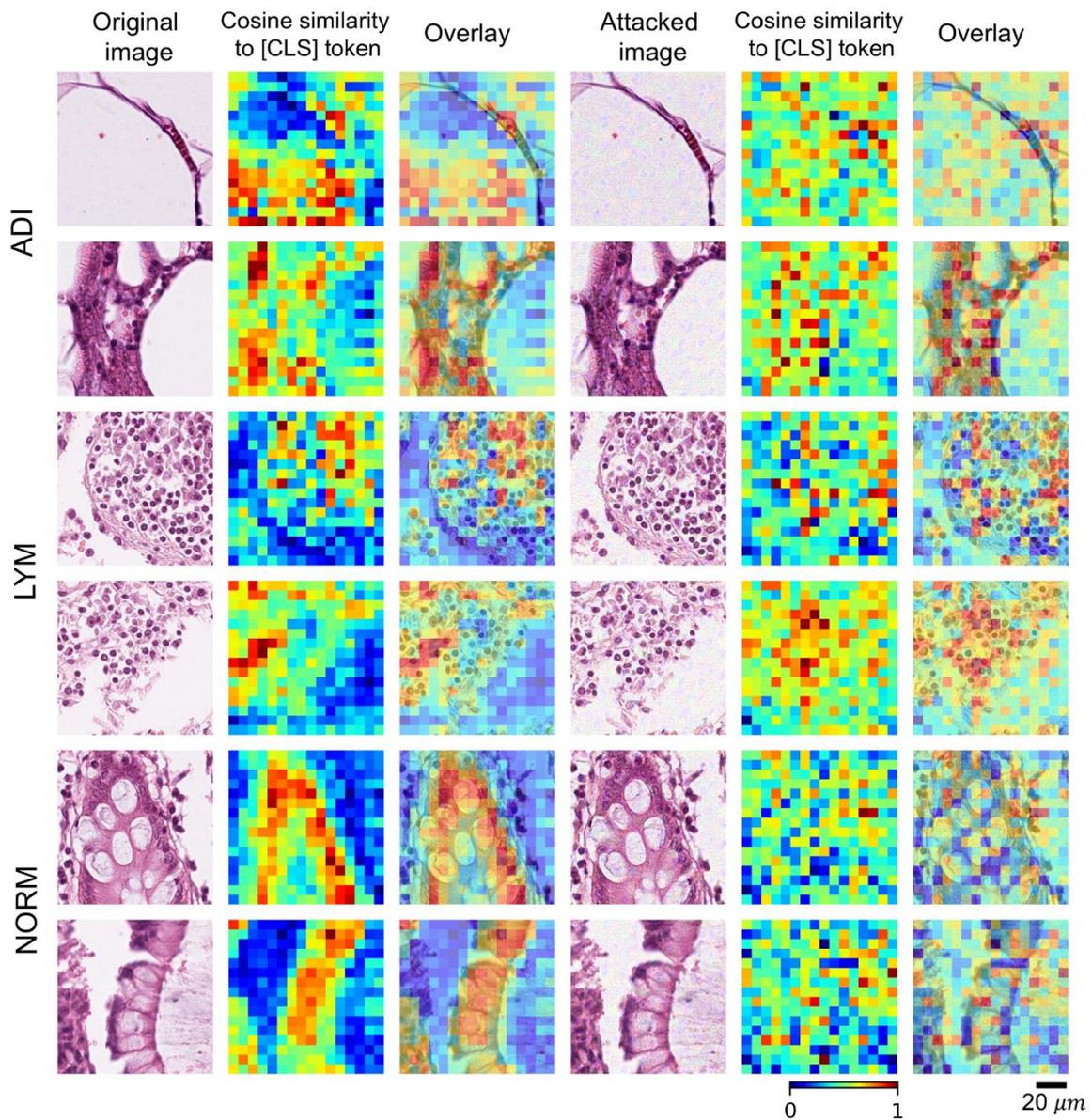

**Figure 3. Original and UTAP attacked attention maps extracted by a pathology foundation model (UNI2-h).** Visualization of the cosine similarity between the [CLS] token and all positional patch tokens for tissue samples from three different classes (ADI, LYM, and NORM). For the original images (left column), the foundation model exhibits coherent attention, with high similarity (red color) correctly localized on salient tissue structures, as desired and expected. After applying the UTAP attack (right column), these attention patterns result in disorganized and scattered states. This provides direct evidence

that UTAP fundamentally corrupts the foundation model's core feature extraction capabilities. The attention maps are calculated using UNI2-h model.

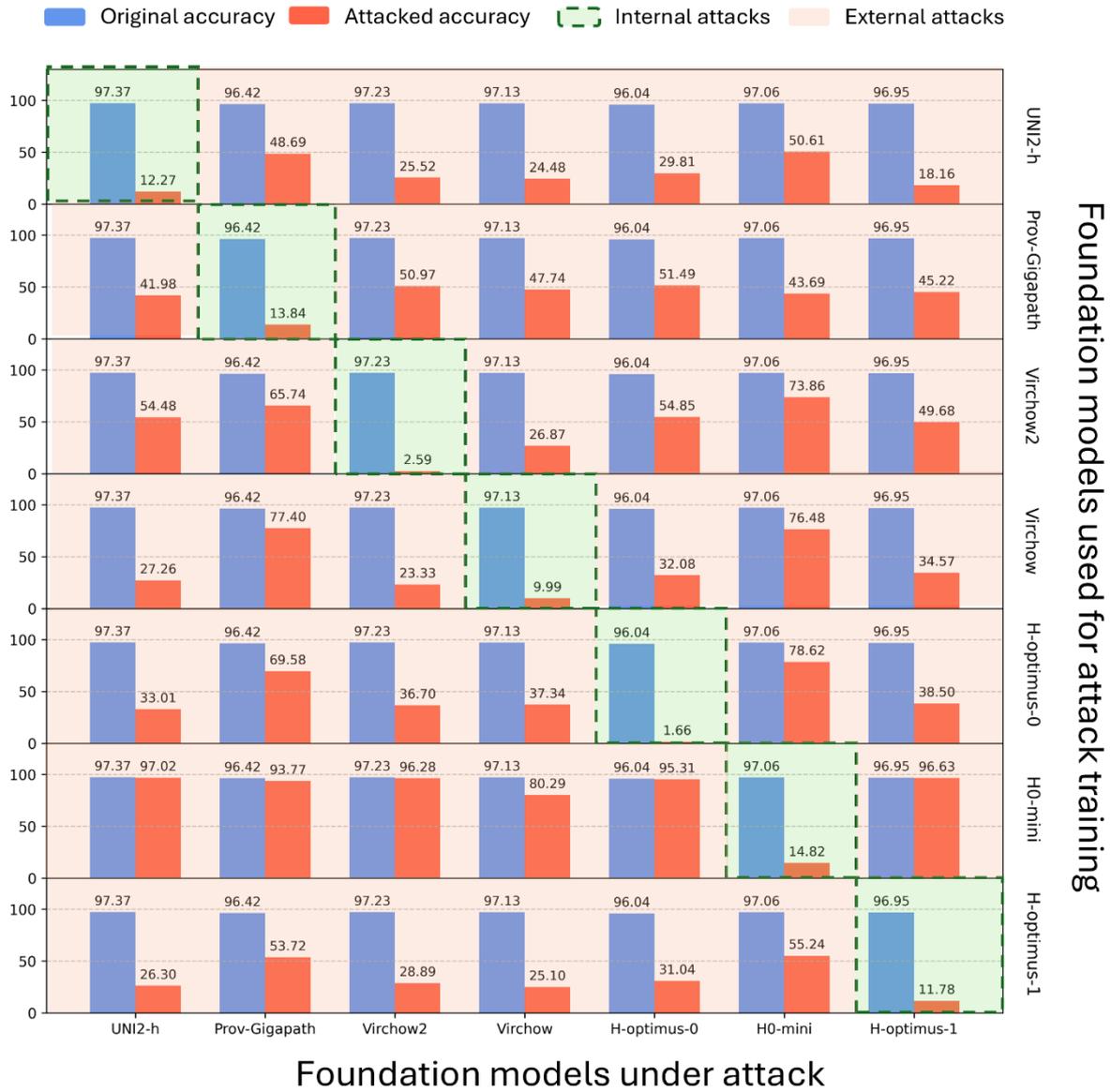

**Figure 4. Cross-model transferability of UTAP.** The matrix displays the attack performance of UTAP perturbations trained on each foundation model (rows) when evaluated against all the other foundation models in the test set (columns). The diagonal elements, highlighted by the green dashed boxes, represent the internal attacks, while the off-diagonal elements represent the external attacks on new foundation models never seen before. In each subplot, the blue bar indicates the original accuracy of the target model, and the red bar shows the accuracy after the UTAP attack.

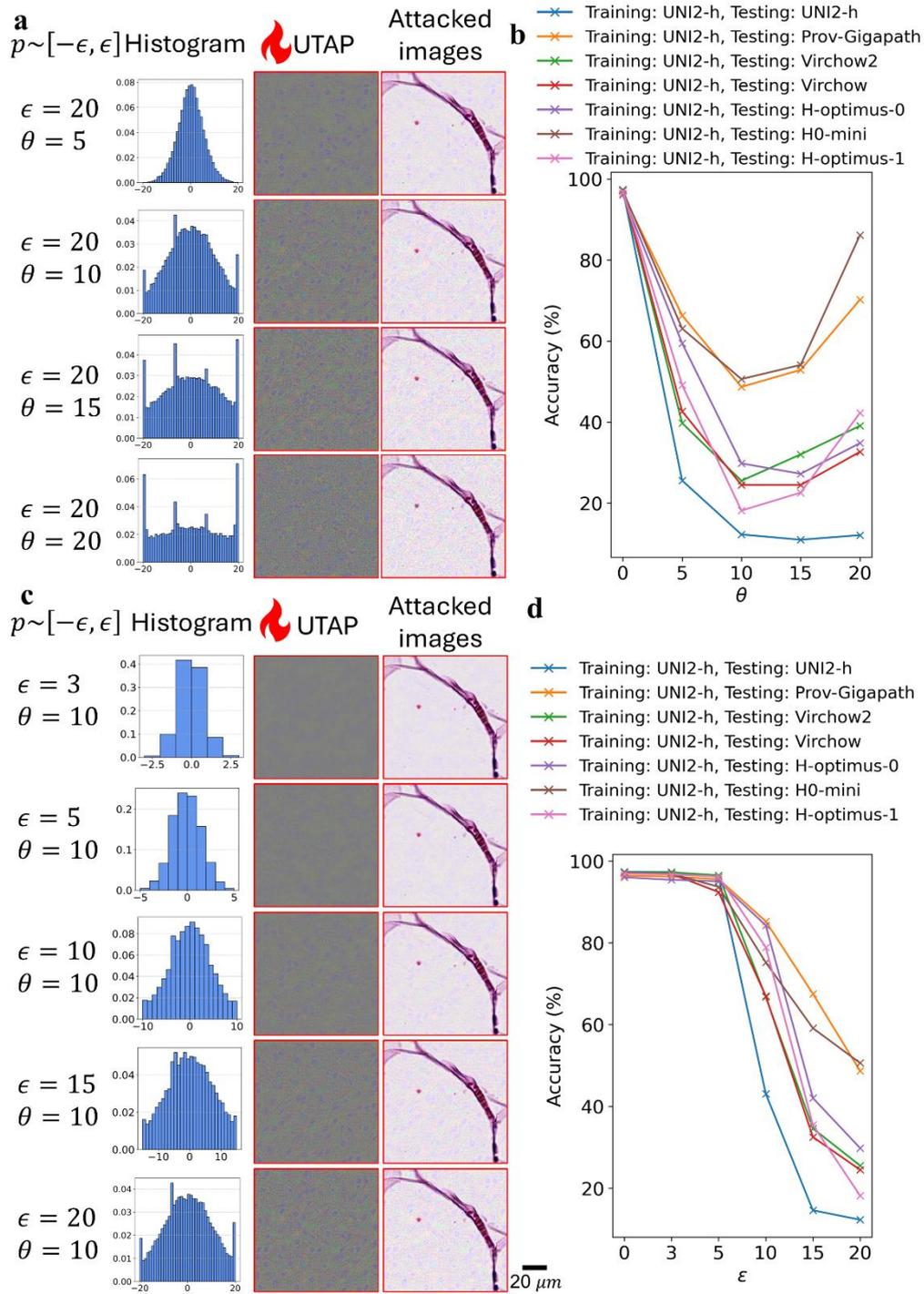

**Figure 5. Ablation study on UTAP hyperparameters $\theta$ and $\epsilon$.** (a) Visualization of UTAP attack with a fixed perturbation magnitude $\epsilon$ and varying step-size scaling factor $\theta$. Each row shows the resulting perturbation histogram, UTAP visualization, and the attacked image. (b) The classification accuracy of seven foundation models (including internal and external attacks) as a function of $\theta$. (c) Visualization of UTAP attacks with a fixed $\theta$ and varying $\epsilon$. (d) The classification accuracy of seven foundation models (including internal and external attacks) as a function of $\epsilon$.

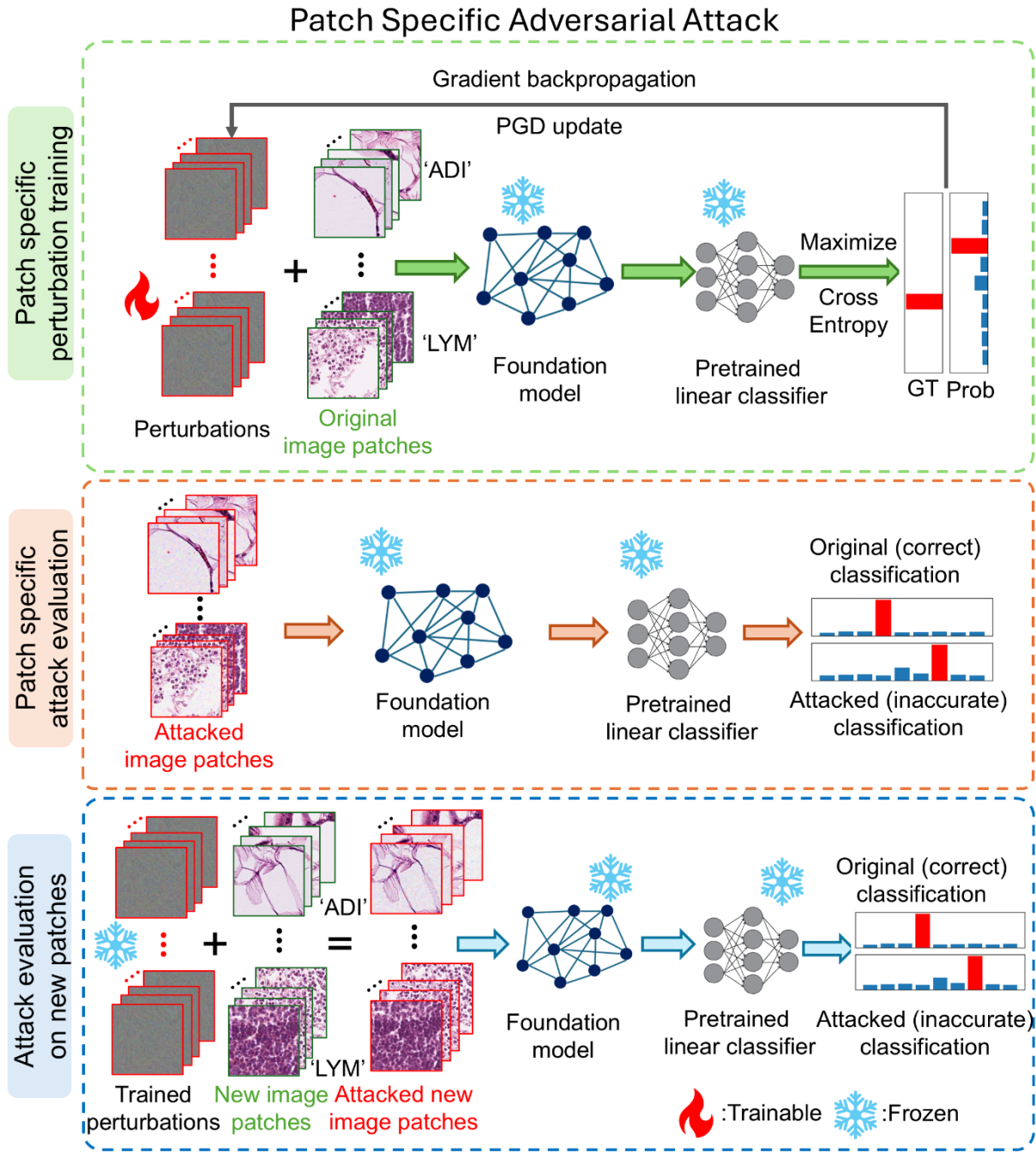

**Figure 6. Workflow of Patch Specific Adversarial Perturbation (PSAP) training and evaluation.** The top row shows the process of PSAP training, where the patch-specific perturbations were added to the original image patches and were optimized to maximize the cross entropy of the ground truth label and the classification probability extracted from the pretrained frozen linear classifier. The remaining rows show

the process of PSAP evaluation on trained (middle row) and new (bottom row) image patches. The fire and snowflake symbols indicate whether the parameters are trainable or frozen/fixed.

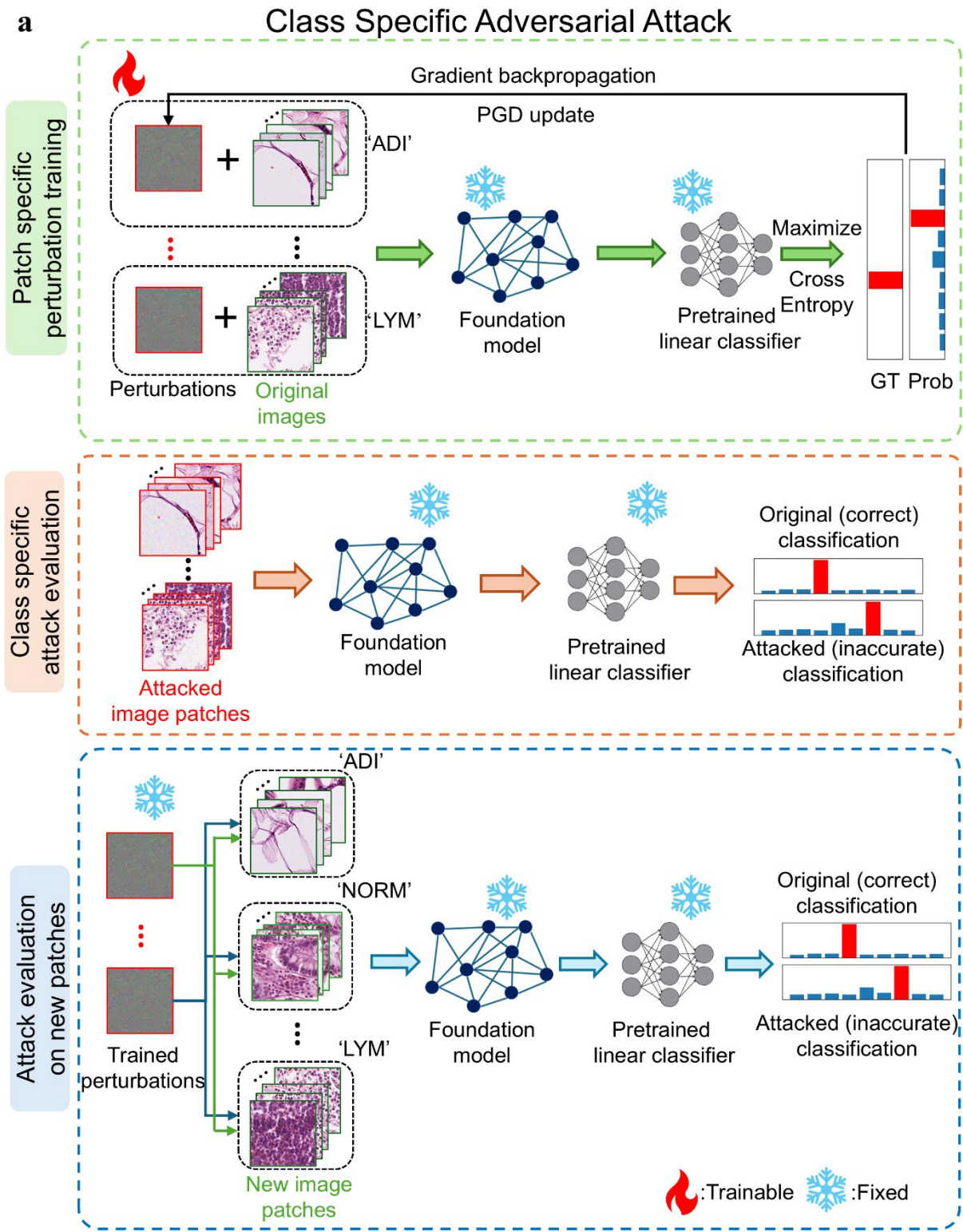

**Figure 7. Workflow of Class Specific Adversarial Perturbation (CSAP) training and evaluation.** The top row shows the process of CSAP training, where the class-specific perturbations were added to the original image patches of a specific data class and were optimized to maximize the cross entropy of the ground truth label and the classification probability extracted from the pretrained frozen linear classifier. The remaining rows show the process of CSAP evaluation on trained (middle row) and new (bottom row) image patches. The fire and snowflake symbols indicate whether the parameters are trainable or frozen/fixed.

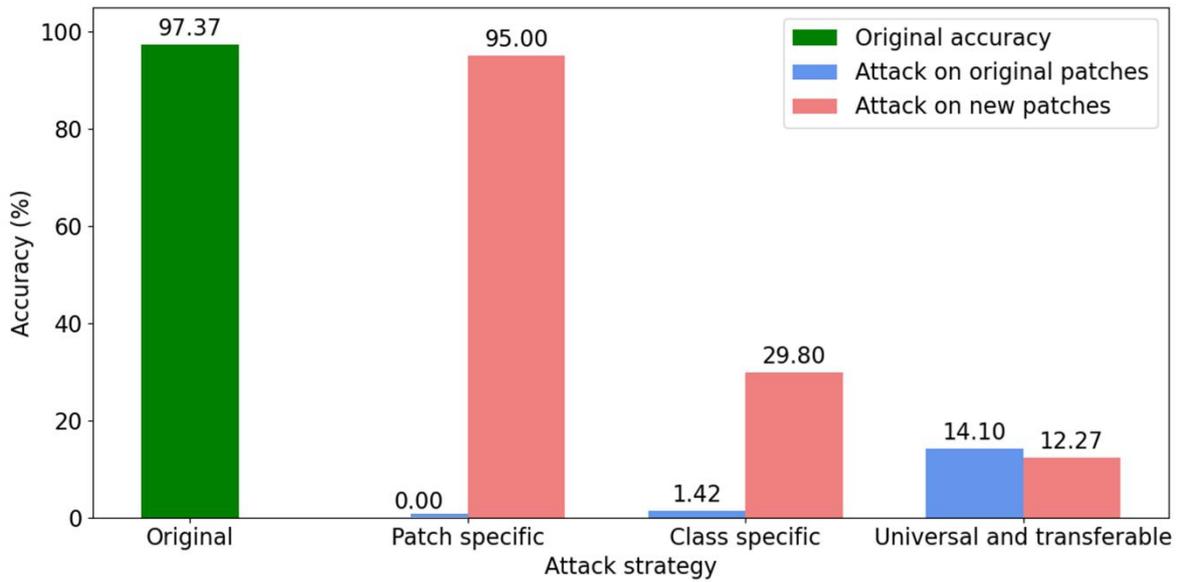

**Figure 8. Comparison of adversarial attack strategies and their generalization.** The green bar indicates the original classification accuracy. The classification accuracies were evaluated on CRC-100K dataset using the UNI2-h foundation model. The attacked classification accuracy values on the original/training and new/testing image patches are shown with the blue and red bars, respectively. The methods of applying these three categories of perturbations are shown in **Fig. 1**, **Fig. 6** and **Fig. 7**.